\documentclass{llncs}
\usepackage{makeidx}
\usepackage[english]{babel}
\usepackage[utf8]{inputenc}
\usepackage{amsmath}
\usepackage{url}
\usepackage{fancyvrb}
\usepackage[spanish]{babelbib}
\usepackage{graphicx}

\usepackage{tikz}
\usetikzlibrary{shadows}

\graphicspath{ {img/} }

\urldef{\mails}\path|{fbarrios,fjlopez}@fi.uba.ar| 

\urldef{\mails}\path|{fbarrios,fjlopez}@fi.uba.ar| 

\spnewtheorem{definicion}[theorem]{Definition}{\bfseries}{\itshape}

\begin{document}

\frontmatter

\title{Variations of the Similarity Function of TextRank for Automated Summarization}
\titlerunning{Variantes de TextRank} 

\author{Federico Barrios\inst{1}, Federico López\inst{1}, Luis Argerich\inst{1}, Rosita Wachenchauzer\inst{1}\inst{2}}
\institute{Facultad de Ingeniería, Universidad de Buenos Aires,\\Ciudad Autónoma de Buenos Aires, Argentina. \\
\and Universidad Nacional de Tres de Febrero, Caseros, Argentina.\\
\mails}

\maketitle

\begin{abstract}
This article presents new alternatives to the similarity function for the TextRank algorithm for automated summarization of texts. We describe the generalities of the algorithm and the different functions we propose. Some of these variants achieve a significative improvement using the same metrics and dataset as the original publication. 

\end{abstract}

\section{Introduction}
In the field of natural language processing, an extractive summarization task can be described as the selection of the most important sentences in a document. Using different levels of compression, a summarized version of the document of arbitrary length can be obtained.

TextRank is a graph-based extractive summarization algorithm. It is domain and language independent since it does not require deep linguistic knowledge, nor domain or language specific annotated corpora \cite{mihalcea}. These features makes the algorithm widely used: it performs well summarizing structured text like news articles, but it has also shown good results in other usages such as summarizing meeting transcriptions \cite{garg} and assessing web content credibility \cite{balcerzak}.

In this article we present different proposals for the construction of the graph and report the results obtained with them.

The first sections of this article describe previous work in the area and an overview of the TextRank algorithm. Then we present the variations and describe the different metrics and dataset used for the evaluation. Finally we report the results obtained using the proposed changes.

\section{Previous work}
The field of automated summarization has attracted interest since the late 50's \cite{luhn}. Traditional methods for text summarization analyze the frequency of words or sentences in the first paragraphs of the text to identify the most important lexical elements. The mainstream research in this field emphasizes extractive approaches to summarization using statistical methods \cite{a_survey}. Several statistical models have been developed based on training corpora to combine different heuristics using keywords, position and length of sentences, word frequency or titles \cite{hovy}. 

Other methods are based in the representation of the text as a graph. The graph-based ranking approaches consider the intrinsic structure of the texts instead of treating texts as simple aggregations of terms. Thus it is able to capture and express richer information in determining important concepts \cite{similarity_functions}.

The selected text fragments to use in the graph construction can be phrases \cite{lexical_chains}, sentences \cite{luhn}, or paragraphs \cite{scalable_summarization}. Currently, many successful systems adopt the sentences considering the tradeoff between content richness and grammar correctness. According to these approach the most important sentences are the most connected ones in the graph and are used for building a final summary \cite{barzilay}. To identify relations between sentences (edges for the graph) there are several measures: overlapping words, cosine distance and query-sensitive similarity. Also, some authors have proposed combinations of the previous with supervised learning functions \cite{similarity_functions}.

These algorithms use different information retrieval techniques to determine the most important sentences (vertices) and build the summary \cite{salton}. The TextRank algorithm developed by Mihalcea and Tarau \cite{mihalcea-tarau} and the LexRank algorithm by Erkan and Radev \cite{erkan} are based in ranking the lexical units of the text (sentences or words) using variations of PageRank \cite{pageetal98}. Other graph-based ranking algorithms such as HITS \cite{kleinberg} or Positional Function \cite{herings} may be also applied.

\section{TextRank}

\subsection{Description}
TextRank is an unsupervised algorithm for the automated summarization of texts that can also be used to obtain the most important keywords in a document. It was introduced by Rada Mihalcea and Paul Tarau in \cite{mihalcea-tarau}.

The algorithm applies a variation of PageRank \cite{pageetal98} over a graph constructed specifically for the task of summarization. This produces a ranking of the elements in the graph: the most important elements are the ones that better describe the text. This approach allows TextRank to build summaries without the need of a training corpus or labeling and allows the use of the algorithm with different languages.

\subsection{Text as a Graph}
For the task of automated summarization, TextRank models any document as a graph using sentences as nodes \cite{introductionir}. A function to compute the similarity of sentences is needed to build edges in between. This function is used to weight the graph edges, the higher the similarity between sentences the more important the edge between them will be in the graph. In the domain of a Random Walker, as used frequently in PageRank \cite{pageetal98}, we can say that we are more likely to go from one sentence to another if they are very similar. 

TextRank determines the relation of similarity between two sentences based on the content that both share. This overlap is calculated simply as the number of common lexical tokens between them, divided by the lenght of each to avoid promoting long sentences.

The function featured in the original algorithm can be formalized as:

\begin{definicion}
Given $S_i$, $S_j$ two sentences represented by a set of $n$ words that in 
$S_i$ are represented as $S_i = w_{1}^{i}, w_{2}^{i},..., w_{n}^{i}$. The similarity function for $S_i$, $S_j$ can be defined as:

\begin{equation}
Sim(S_{i},S_{j}) = \frac{ | \{   w_{k} | w_{k} \in S_{i} \& w_{k} \in S_{j}   \}  | }    
                              {  log(|S_{i}|) + log(|S_{j}|)  }
\end{equation}

\end{definicion}

The result of this process is a dense graph representing the document. From this graph, PageRank is used to compute the importance of each vertex. The most significative sentences are selected and presented in the same order as they appear in the document as the summary.

\section{Experiments}

\subsection{Our Variations}
This section will describe the different modifications that we propose over the original TextRank algorithm. These ideas are based in changing the way in which distances between sentences are computed to weight the edges of the graph used for PageRank. These similarity measures are orthogonal to the TextRank model, thus they can be easily integrated into the algorithm. We found some of these variations to produce significative improvements over the original algorithm.

\subsubsection{Longest Common Substring}
From two sentences we identify the longest common substring and report the similarity to be its length \cite{gusfield}.

\subsubsection{Cosine Distance}
The cosine similarity is a metric widely used to compare texts represented as vectors. We used a classical TF-IDF model to represent the documents as vectors and computed the cosine between vectors as a measure of similarity. Since the vectors are defined to be positive, the cosine results in values in the range [0,1] where a value of 1 represents identical vectors and 0 represents orthogonal vectors \cite{singhal}.

\subsubsection{BM25}
BM25 / Okapi-BM25 is a ranking function widely used as the state of the art for Information Retrieval tasks. BM25 is a variation of the TF-IDF model using a probabilistic model \cite{robertson}.

\begin{definicion}
Given two sentences R, S, BM25 is defined as:

\begin{equation}
BM25(R,S) = \sum_{i=1}^{n} IDF(s_i) \cdot \frac{f(s_i, R) \cdot (k_1 + 1)}{f(s_i, R) + k_1 \cdot (1 - b + b \cdot \frac{|R|}{avgDL})}
\end{equation}

where $k$ and $b$ are parameters. We used $k = 1.2$ and $b = 0.75$. $avgDL$ is the average length of the sentences in our collection.
\end{definicion}

This function definition implies that if a word appears in more than half the documents of the collection, it will have a negative value. Since this can cause problems in the next stage of the algorithm, we used the following correction formula:
                
\begin{equation}
 IDF(s_i) =
  \begin{cases}
       log(N - n(s_i) + 0.5) - log(n(s_i) + 0.5)    & \text{if }  n(s_i) > N/2\\
       \varepsilon \cdot avgIDF                     & \text{if }  n(s_i) \leq N/2\\
  \end{cases}
\end{equation}                
                
where $\varepsilon$ takes a value between 0.5 and 0.30 and $avgIDF$ is the average IDF for all terms.
Other corrective strategies were also tested, setting $\varepsilon$ = 0 and using simpler modifications of the classic IDF formula.

We also used BM25+, a variation of BM25 that changes the way long documents are penalized \cite{lv}.

\subsection{Evaluation}
For testing the proposed variations, we used the database of the 2002 Document Understanding Conference (DUC) \cite{duc2002-guidelines}. The corpus has 567 documents that are summarized to 20\% of their size, and is the same corpus used in \cite{mihalcea-tarau}. 

To evaluate results we used version 1.5.5 of the ROUGE package \cite{Lin2004a}. The configuration settings were the same as those in DUC, where ROUGE-1, ROUGE-2 and ROUGE-SU4 were used as metrics, using a confidence level of 95\% and applying stemming. The final result is an average of these three scores.

To check the correct behaviour of our test suite we implemented the reference method used in \cite{mihalcea-tarau}, which extracts the first sentences of each document. We found the resulting scores of the original algorithm to be identical to those reported in \cite{mihalcea-tarau}: a 2.3\% improvement over the baseline.

\subsection{Results}
We tested LCS, Cosine Sim, BM25 and BM25+ as different ways to weight the edges for the TextRank graph. 
The best results were obtained using BM25 and BM25+ with the corrective formula shown in equation 3. We achieved an improvement of 2.92\% above the original TextRank result using BM25 and \mbox{$\varepsilon$ = 0.25}. The following chart shows the results obtained for the different variations we proposed.

\begin{table}
\caption{Evaluation results for the proposed TextRank variations.}
\begin{center}
\begin{tabular}{l*{5}{c}r}
\hline
\rule{0pt}{12pt}
Method & ROUGE-1 & ROUGE-2 & ROUGE-SU4 & Improvement \\[2pt]
\hline\rule{0pt}{12pt}\mbox{}\par\nobreak
BM25 ($\varepsilon$ = 0.25) & 0.4042 & 0.1831 & 0.2018 & 2.92\% \\
BM25+ ($\varepsilon$ = 0.25) & 0.404 & 0.1818 & 0.2008 & 2.60\% \\
Cosine TF-IDF & 0.4108 & 0.177 & 0.1984 & 2.54\% \\
BM25+ (IDF = log($N$/$N_i$)) & 0.4022 & 0.1805 & 0.1997 & 2.05\% \\ 
BM25 (IDF = log($N$/$N_i$)) & 0.4012 & 0.1808 & 0.1998 & 1.97\% \\ 
Longest Common Substring & 0.402 & 0.1783 & 0.1971 & 1.40\% \\
BM25+ ($\varepsilon$ = 0) & 0.3992 & 0.1803 & 0.1976 & 1.36\% \\ 
BM25 ($\varepsilon$ = 0) & 0.3991 & 0.1778 & 0.1966 & 0.89\% \\
\textbf{TextRank} & \textbf{0.3983} & \textbf{0.1762} & \textbf{0.1948} & \textbf{--}\\
BM25 & 0.3916 & 0.1725 & 0.1906 & -1.57\% \\
BM25+ & 0.3903 & 0.1711 & 0.1894 & -2.07\% \\
DUC Baseline & 0.39 & 0.1689 & 0.186 & -2.84\% \\ [2pt]
\hline
\end{tabular}
\end{center}
\end{table}

\begin{figure}[h!]
    \centering
    \includegraphics[width=1\textwidth]{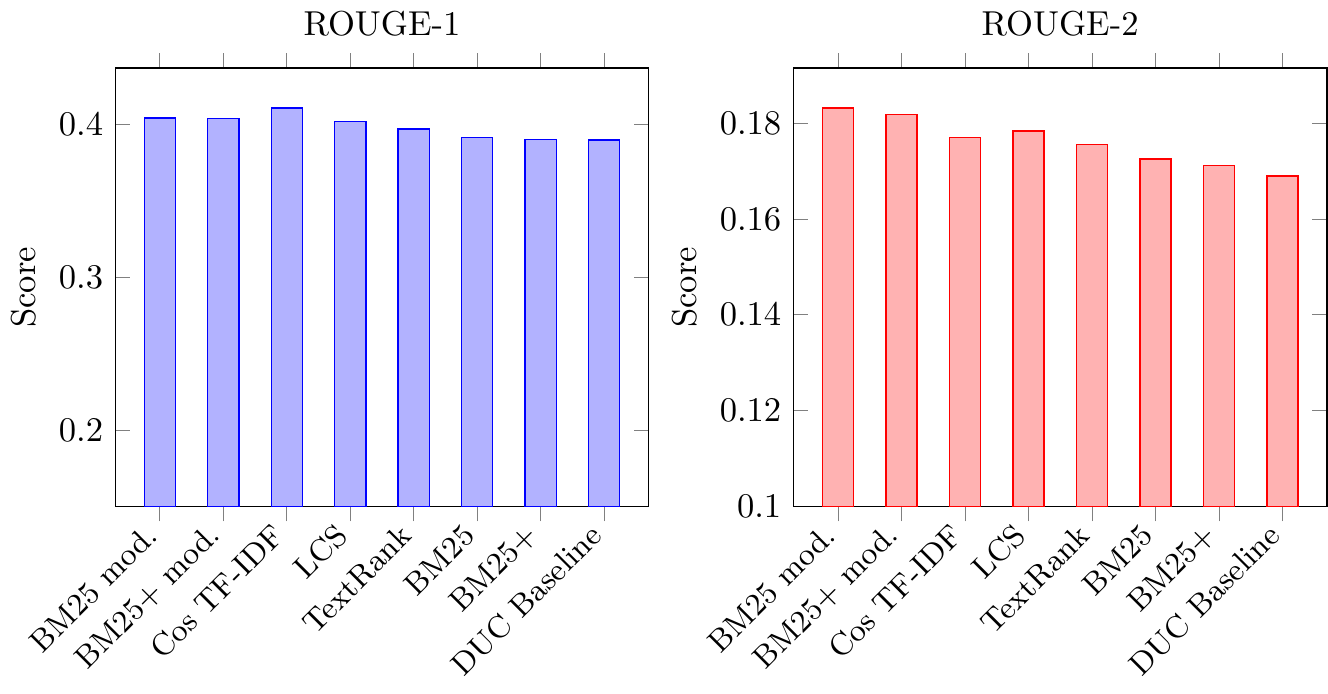}
    \caption{ROUGE-1 and ROUGE-2 scores comparison.}
\end{figure}

The result of Cosine Similarity was also satisfactory with a 2.54\% improvement over the original method. The LCS variation also performed better than the original TextRank algorithm with 1.40\% total improvement.

The performance in time was also improved. We could process the 567 documents from the DUC2002 database in 84\% of the time needed in the original version.

\section{Reference Implementation and Gensim Contribution}
A reference implementation of our proposals was coded as a Python module\footnote{Source code available at: \url{https://github.com/summanlp/textrank}} and can be obtained for testing and to reproduce results.

We also contributed the BM25-TextRank algorithm to the Gensim project\footnote{Source code available at: \url{https://github.com/summanlp/gensim}} \cite{rehurek_lrec}.

\section{Conclusions}
This work presented three different variations to the TextRank algorithm for automatic summarization. The three alternatives improved significantly the results of the algorithm using the same test configuration as in the original publication. Given that TextRank performs 2.84\% over the baseline, our improvement of 2.92\% over the TextRank score is an important result. 

The combination of TextRank with modern Information Retrieval ranking functions such as BM25 and BM25+ creates a robust method for automatic summarization that performs better than the standard techniques used previously. 

Based on these results we suggest the use of BM25 along with TextRank for the task of unsupervised automatic summarization of texts. The results obtained and the examples analyzed show that this variation is better than the original TextRank algorithm without a performance penalty.

\bibliography{articulo-en}{}
\bibliographystyle{splncs03}

\end{document}